\documentclass[conference]{IEEEtran}
\IEEEoverridecommandlockouts
\usepackage{cite}
\usepackage{amsmath,amssymb,amsfonts}
\usepackage{algorithmic}
\usepackage{graphicx}
\usepackage{textcomp}
\usepackage{xcolor}
\usepackage{tabularx,booktabs}
\usepackage{comment} 
\PassOptionsToPackage{hyphens}{url}\usepackage{hyperref}

\usepackage{amsmath,amssymb,amsfonts}
\usepackage{algorithmic}
\usepackage{graphicx}
\usepackage{subfiles}
\usepackage{textcomp}
\usepackage{xcolor}

\newtheorem{definition}{Definition}
\newtheorem{problem}{Problem}

\def\BibTeX{{\rm B\kern-.05em{\sc i\kern-.025em b}\kern-.08em
    T\kern-.1667em\lower.7ex\hbox{E}\kern-.125emX}}
\begin{document}

\title{Atypical lexical abbreviations identification in Russian medical texts\\}

\author{\IEEEauthorblockN{Anna Berdichevskaia}
\IEEEauthorblockA{\textit{Faculty of Engineering Cybernetics} \\
\textit{NUST MISiS}\\
Moscow, Russia \\
berdichevskaya.a.g@gmail.com}
}

\maketitle


\begin{abstract}
Abbreviation is a method of word formation that aims to construct the shortened term from the first letters of the initial phrase. Implicit abbreviations frequently cause the comprehension difficulties for unprepared readers. In this paper, we propose an efficient ML-based algorithm which allows to identify the abbreviations in Russian texts. The method achieves ROC AUC score 0.926 and F1 score 0.706 which are confirmed as competitive in comparison with the baselines. Along with the pipeline, we also establish first to our knowledge Russian dataset that is relevant for the desired task.

\end{abstract}

\begin{IEEEkeywords}
natural language processing, machine learning, medical texts analysis, abbreviation expansion
\end{IEEEkeywords}

\section{Introduction}
Low awareness of the population in the issues of evidence-based medicine is one of the reasons that encourage people to rely on self-diagnosis and self-medication. Such a condition can directly lead to serious side effects and accidents. One of the steps towards solving this problem is a simplification of access to modern healthcare information. It is important to make medical texts readable for people who do not have specific knowledge in a particular field of medicine.

One of the reasons why medical texts are difficult for non-professionals to understand is widely used lexical abbreviations~\cite{Liu2001}. The average ratio of abbreviations makes up about 7\% of words in medical texts.

Almost a third of the registered medical abbreviations have different meanings depending on the context. For example, "ALT" can be a short form of "alanine transaminase", "argon laser trabeculoplasty" or "alternative lengthening of telomeres". Some tokens (e.g., "WHO") could be interpreted either as an abbreviation (for "World Health Organization" in this case) or an independent word. Another serious obstacle to the coherent comprehension of the medical sources is the widespread use of \textit{ad-hoc} abbreviations.

Some researchers have previously proposed approaches to solve the problem of recognizing atypical abbreviations in medical texts, for example~\cite{Schwartz_and_Hearst} and~\cite{Nadeau2008}. However, these studies were focused on abbreviations in English texts. In~\cite{glass2017language} authors consider a multilingual model which was tested on the Russian texts. The results of this study allow us to conclude that the algorithm can be successfully applied to solve the introduced problem. However, it is possible to get even better results using more suitable data for training a model and considering the specific features of the Russian language.

This work is aimed to identify abbreviations in the text and map them with the possible definitions in a given context. The contributions of this study are the following: 
\begin{enumerate}
    \item We have published a dataset based on the texts of articles from the journal "Problems of Endocrinology". Established materials represent the unique dataset of high novelty according to the absence of counterparts in the case of the Russian language.
    
    \item A new approach to solving the problem of recognizing atypical abbreviations in medical texts was proposed. The approach utilizes two consecutive binary classifications that take as an input custom feature description of sentences obtained via the special procedure of feature engineering.
    
    \item We have done a comprehensive analysis of different model configurations and performed a comparison with existing approaches. As it was demonstrated, the proposed approach tangibly increases performance on the both abbreviation detection and abbreviation identification tasks.
\end{enumerate}

The code and dataset could be accessed via \url{https://github.com/aberdichevskaya/abbreviation_identification}.

\section{Related work}
In this section, we discuss different approaches to abbreviation detection and further definitions search in the text.

\subsection{Abbreviation detection}

Many algorithms leverage heuristic approaches to find a set of potential abbreviations from a given text. Common rules for abbreviation detection include restrictions on abbreviation length, number and percentage of non-letter digits and capital letters~\cite{Terada2004, Nadeau2008, Ciosici2019, Moon2014}. In another study~\cite{Schwartz_and_Hearst} authors use the sign of parentheses as an input feature.

Along with deterministic methods, machine learning also has proven itself as an efficient approach to abbreviations detection. Common features for classification whether a token is an abbreviation or not are the peculiarity of word formation, number and percentage of vowels and consonants, features derived from knowledge bases (as well as from the basic comprehensive corpus), and features derived from the context of a word~\cite{Rosenbloom2011, Toole2000, Xu2007}.

\subsection{Search for potential definitions in the text}

The set of potential definitions for each abbreviation composed of all possible sequences of words taken from the text. Therefore, before applying matching algorithm, authors reduce the space of expected definition with the help of certain heuristics.

In paper~\cite{Pustejovsky} authors introduce the assumption that the first word of the definition should contain the first letter of the abbreviation. 
Other introduced rules~\cite{Park_Byrd} imply limitation of maximum length for a definition. As it follows from the same study, all words in each definition also should be in the same sentence. Additionally, it imposes the restrictions on the part of the speech of the first and the last words in the definition. 
Authors of the study~\cite{Nadeau2008} offered further rules: definition may not contain one letter of the abbreviation (unless it is only two letters long) as well as digits and punctuation characters. The definition also must not include brackets, colons, semicolons, question marks or exclamation marks. 

\subsection{Matching abbreviations with definitions}

As a result of a definitions generation, we receive the particular abbreviation in form of a pairs set consists of an abbreviations and potential definitions. The further task corresponds to the matching of each abbreviation with one of the given definitions.

The first considered paper~\cite{Pustejovsky} introduces the expression-based method ACROMED based on multi-layered matching algorithm. Despite its simplicity, this approach shows that much greater accuracy can be achieved if determining the window size of the text and identifying the long form of abbreviation are considered as the separate tasks. Further development led researchers to the formalization of the task as a machine learning problem. This group of algorithms could be differentiated in sense of input features. \cite{Nadeau2008} utilizes the distance between the abbreviation and the definition, the number of letters of the abbreviation that match the first letters of the words of the definition, and the sign of parentheses. In~\cite{glass2017language} authors proposed features based on semantic similarity between abbreviation and definition. In contrary to previous studies,~\cite{Vo2018} based on utilizing features of a potential abbreviation (such as a ratio of consonants and vowels in it), features of a potential definition, and term frequency features. 


\subsection{Approaches without definitions extraction}

In some researches, authors propose algorithms of abbreviation identification without extracting of a definition for a particular abbreviation. The LMAAE~\cite{Dua2019} provides the recursive algorithm that divides abbreviation into special blocks with further expansion of generated candidates based on clustering. To choose a suitable for a considering context expansion, authors leverage abbreviation sense disambiguation algorithms.

There are many algorithms based on supervised machine learning, for example~\cite{Jaber2021} and~\cite{Koptient2021}. In study~\cite{Joopudi2018} authors proposed a simple convolutional neural network to solve an abbreviation disambiguation task when in~\cite{Patel2021} authors applied the BERT model for the same challenge.

\section{Preliminaries}
The main object of this study is a lexical abbreviation.

\begin{definition}[Lexical abbreviation]
Let $T=\{t_1, .., t_n\}$ be the text – an ordered set of tokens. Let $D=\{t_l, t_{l+1}, .., t_r\}$, $D \subset T$ – a sequence of tokens from T, further referred as the definition. Then, $A(D, T)$ is a lexical abbreviation of $D$ in the text $T$.
\end{definition}


\begin{problem}[Abbreviations detection]
For a given text T it is required to mark up the tokens included in it for whether they are lexical abbreviations.
\end{problem}

This problem is challenging because of the fact that, depending on the context, token $t_i$ can be either an independent word or an abbreviation.

\begin{problem}[Abbreviations identification]
For a given text T and a set $S=\{A_i, .., A_k\}$ of previously obtained lexical abbreviations it is required to extract all pairs $\langle A_i, D_i \rangle$, where $A_i=A(D_i, T)$.
\end{problem}

The complexity of the abbreviations identification is related to the ambiguous nature of the abbreviations where the same abbreviation $A$ could have different definitions $D_i$.

In the following sections, solutions to these problems will be discussed.

\section{Data}
Due to the fact that the task of atypical lexical abbreviations identification in medical texts is not popular regarding the Russian texts, an existing dataset could not be found. In order to overwhelm such a problem, the authors decided to collect the desired data from the medical peer-reviewed journal "Problems of Endocrinology".
As the basis of the dataset, we used the first two volumes of this journal. Initial data was represented in a form of raw text obtained via the PDFMiner Python package. At the first stage, images metadata, links, notes, and affiliation information were removed from the texts. The given array of symbols was further split into separate tokens where the spaces and dots were interpreted as separators. Pronouns and prepositions that did not carry any semantic significance were deleted. Finally, we performed a manual marking of each word in order to determine whether the word is an abbreviation or not.

The dataset contains 16237 tokens and includes 6\% (960) of abbreviations.

\section{Method}
\begin{figure*}[htbp]
\centerline{\includegraphics[width=\textwidth]{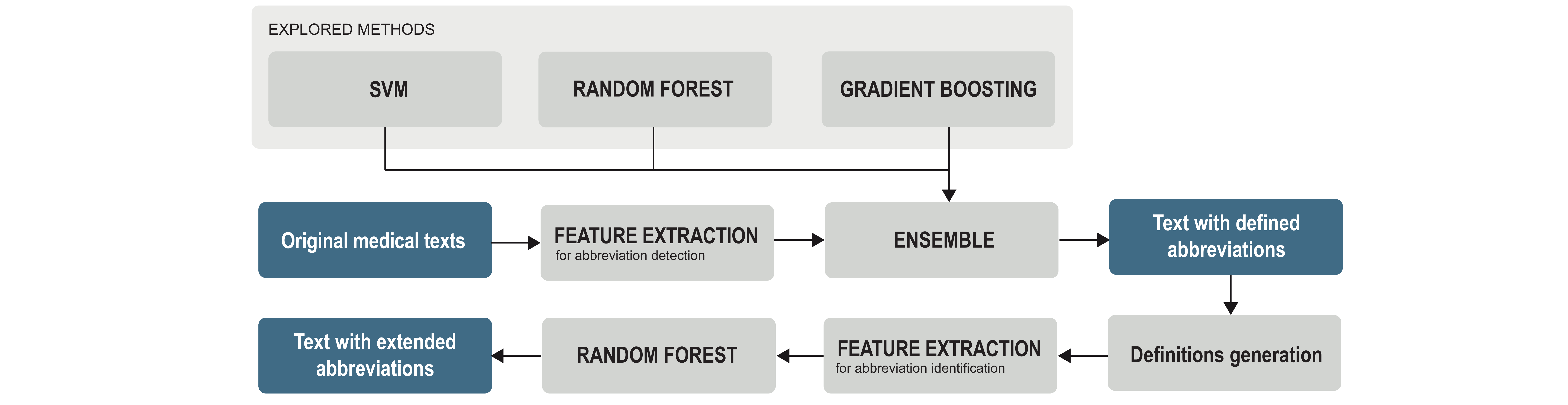}}
\caption{The proposed algorithm consists of five main blocks: feature extractor for an abbreviation detection task, classification of abbreviations, generation of potential definitions, feature extraction for an abbreviation identification task, and classification of correct pairs consist of an abbreviation and its definition.}
\label{pipeline}
\end{figure*}

The proposed method of identifying abbreviations in the text consists of two modules. The first part of the algorithm searches for abbreviations in the text. The obtained output is fed into the second module that extracts a set of definitions for each found abbreviation and predicts its exact form. Visual description of the pipeline is shown in Figure~\ref{pipeline}.

\subsection{Abbreviation detection}

At the first stage, words are checked for belonging to the class of abbreviations via one of the considered machine learning algorithms. Formally, the task of detecting abbreviations in the text is reduced to determining by a given feature vector $X=\{x_1,x_2,x_3,x_4,x_5,x_6\}$ if a token $t$ belongs to a class of full words or a class of lexical abbreviations. Following features are used as input for the binary classifier:

\begin{itemize}
    \item Presence of any special characters (such as dots, hyphens, slashes, and so on) in the token. 
    \item Presence of only letters, only numbers, or their combination. 
    \item Presence of only vowels, only consonants, or their combination. 
    \item Length of the token.
    \item Percentage of capital letters in the token.
    \item Presence of any capital letters inside the token. This implies that some characters other than the first one are capital letters.
    \item Presence of the token in the Russian dictionary.
\end{itemize}

\subsection{Set of possible definitions of every abbreviation generation}

In this stage previously detected as abbreviation tokens involved as a part of definitions extraction algorithm. It is important to mention that every ordered sequence of tokens from text $T$ may be a definition of a particular abbreviation $A_i \in S$. To restrict the cardinality of this set, the following rules are used:

\begin{itemize}
    \item Length of a definition should not exceed $len_{max}$ that is defined as follows:
    \begin{equation}
        len_{max}=min(length(A_i)+5, length(A_i) \ast 2) \label{eq}.
    \end{equation}
    The length of a definition is equal to the number of tokens in it, and the length of an abbreviation is equal to the number of characters in it.
    \item Let $t_a=A_i, t_a \in T$ and 
    \begin{equation}
        \delta=min(len_{max}+5, 2 \ast len_{max}) \label{eq}.
    \end{equation}
    Then definitions are generated only by sequences $\{t_{a-\delta}, .., t_{a-1}\}$ and $\{t_{a+1}, .., t_{a+\delta}\}$.
    \item The first letter of the abbreviation should be present in the first word of the potential definition. 
    \item At least 80\% of the characters from the abbreviation should be present in the words from the definition.
    \item Maximum one character from the abbreviation may not be present in the definition.
    \item The definition can not contain an abbreviation as a substring.
\end{itemize}

\subsection{Correct pairs extraction}

After generation of possible definitions, the set $M$ of pairs $\langle A_i,D_{j_i}\rangle$ is constructed for every $A_i \in S$. Here $\{D_{j_i}\}$ is a set of all possible definitions for an abbreviation $A_i$. 

 Let $\langle A_w,D_{b_w}\rangle \in M,\;A(D_{b_w}, T)=A_w$ be a correct pair. The next step is to select all of the correct pairs from M. This task can be formulated as a binary classification. As an input of classifier, we utilize a wide range of features: 
\begin{itemize}
    \item The distance between the abbreviation and the definition in the text. If $t_a=A_i, t_{d_l}$ is a first token in $D_{j_i}, t_{d_r}$ is the last token in $D$. Then the distance $\theta$ between abbreviation and the definition can be computed as
    \begin{equation}
        \theta = min(|a-d_l|, |a-d_r|) \label{eq}.
    \end{equation}
    \item The number of letters of the abbreviation that match the first letters of tokens from the definition.
    \item If either an abbreviation or an extension is located inside the parenthesis.
    \item Semantic similarity between abbreviation and definition.
    \item The longest common subsequence of the abbreviation and the definition.
\end{itemize}

Let us take a closer look at the semantic similarity between abbreviation and definition. Consider $E(t)$ as a function that returns embedding of a token t. In this study, the pre-trained RuBERT model~\cite{rubert} was used as the text encoder. Let $\Delta$ be a value defined as
\begin{equation}
    \Delta=\frac{1}{length(D)}*\displaystyle\sum_{t_i \in D} E(t_i). \label{eq}
\end{equation}

The degree of synonymy between abbreviation A and definition D is defined as
\begin{equation}
    H = sim_{cos}(E(A), \Delta). \label{eq}
\end{equation}

\section{Results}

\begin{table*}[t]

\begin{center}
\renewcommand{\arraystretch}{1.5}

\begin{minipage}[t]{.5\linewidth}
    \centering
    
    \caption{Evaluation of different approaches for \\ abbreviation detection}
    \label{tab:detection_results}

    \medskip
    
\begin{tabularx}{\textwidth}{|c|X|X|X|}

\hline

\textbf{Model} & \textbf{\textit{ROC AUC}}& \textbf{\textit{Accuracy}}& \textbf{\textit{F1}} \\
\hline

Wu \textit{et al.} (2011) & 0.927 & 0.985 & 0.88 \\
\hline
\hline

SVM & 0.894 & 0.977 & 0.792 \\
\hline

Random Forest & 0.908 & 0.985 & 0.869 \\
\hline

Gradient Boosting & 0.853 & 0.977 & 0.794 \\
\hline

Ensemble & \textbf{0.94} & \textbf{0.99} & \textbf{0.894}\\
\hline

\end{tabularx}
\end{minipage}\hfill
\begin{minipage}[t]{.5\linewidth}
    \centering

    \caption{Evaluation of different approaches for \\ abbreviation identification}
    \label{tab:identification_result}

    \medskip
\begin{tabular}{|c|l|l|l|}
\hline

\textbf{Model} & \textbf{\textit{ROC AUC}}& \textbf{\textit{Accuracy}}& \textbf{\textit{F1}} \\
\hline

Schwartz and Hearst (2003) & 0.788 & 0.627 & 0.692 \\
\hline

Glass \textit{et al.} (2017) & 0.918 & 0.761 & 0.701 \\
\hline
\hline

SVM & 0.924 & 0.99 & 0.6 \\
\hline

Gradient Boosting & \textbf{0.926} & \textbf{0.994} & \textbf{0.706} \\
\hline

Random Forest & \textbf{0.926} & \textbf{0.994} & \textbf{0.706} \\
\hline

\end{tabular}
\end{minipage}\hfill
\label{tab1}
\end{center}
\end{table*}

\begin{figure}[t]
\centerline{\includegraphics[scale=0.19]{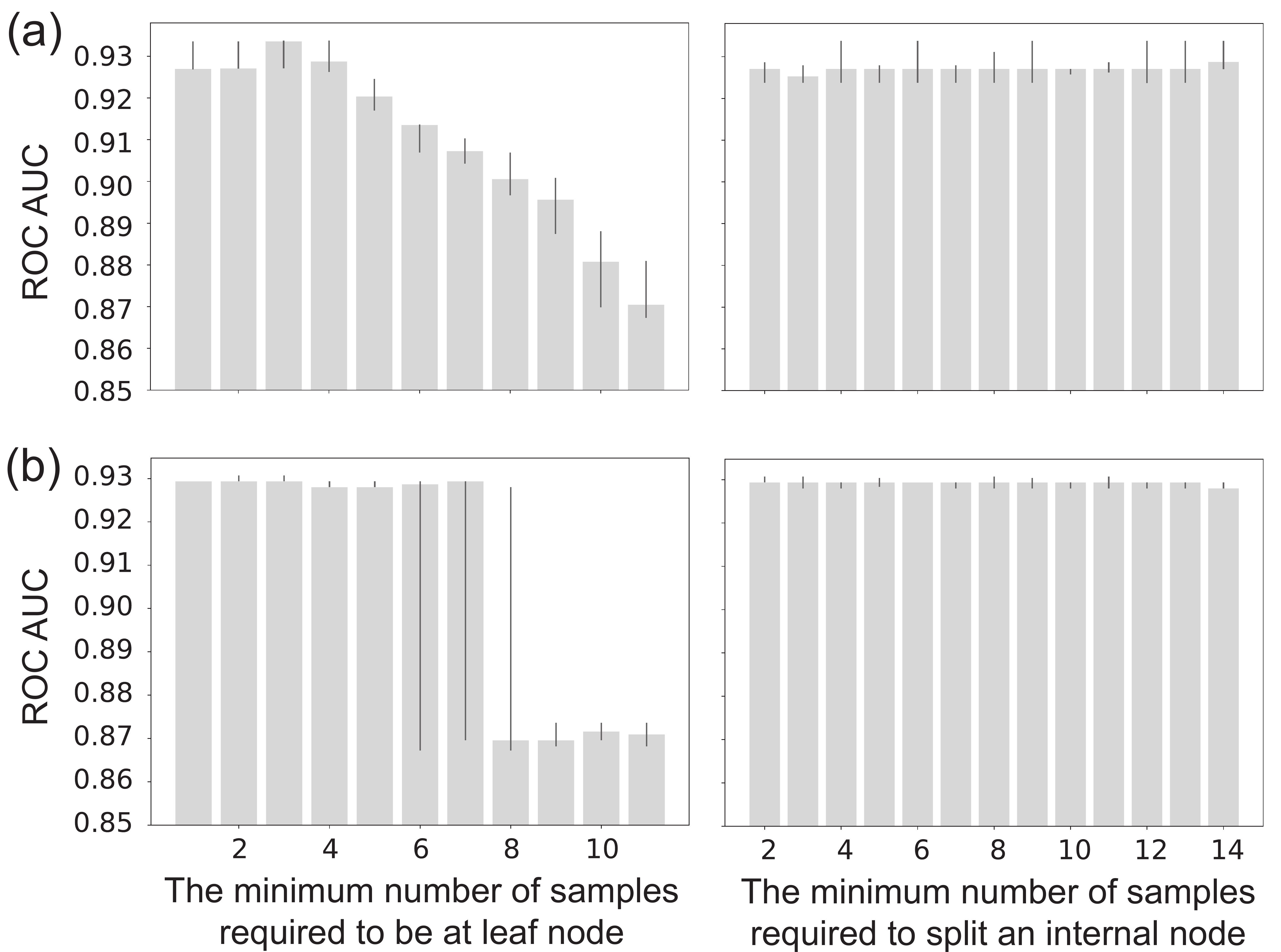}}
\caption{Influence of model parameters on the ROC AUC score: (a) refers to the abbreviation detection task and (b) refers to the abbreviation identification.}
\label{parametrs}
\end{figure}

In this section, we compare the results of our computational experiments and examine the dependencies of the model on parameters. 

For all experiments, the dataset was randomly divided into train and test samples in a ratio 4:1. To measure the performance of models, we used ROC AUC, accuracy, and F1 metrics.


\subsection{Abbreviation detection}

During the experiments, we utilized Support Vector Machine (SVM), Random Forest, and Gradient Boosting. 

In the algorithm based on SVM, each token is represented as a point in $n$-dimensional space. The algorithm divides points into two classes via hyperplane so that the distance between the hyperplane and the nearest point from any group is maximized.

The random forest algorithm is a committee of decision trees. Let us define $r=\sqrt{n}$ where $n$ is the dimension of the feature space. Each tree classifies a random subsample of elements based on $r$ randomly chosen features. The classification of sample objects is performed by voting between the trees. The prediction is a class for which the majority of trees voted.

Gradient boosting is a committee of weak classifiers such as decision trees. The final classification of the sample objects corresponds to a linear combination of the results of the weak classifiers and the coefficient that is inversely proportional to the error of the classifier on the training sample.
 
As an addition to the previously mentioned models, we propose to combine them into the ensemble of classifiers. The voting scheme of the ensemble is the following: if at least one of the algorithms has assumed a token as an abbreviation, then this token is considered further as an abbreviation.

In comparison with other models, the ensemble shows the best performance with ROC AUC score \textbf{0.94}, accuracy \textbf{0.99}, and F1-score \textbf{0.894}. From 4249 tokens in the test sample, an algorithm correctly found 214 abbreviations, incorrectly mark 25 tokens as abbreviations and 37 abbreviations were overlooked.

In order to assess the proposed algorithm objectively, we compared it with the actual baseline~\cite{Rosenbloom2011}, Table~\ref{tab:detection_results}. The ensemble showed a notable improvement of detection quality which proves the perspective of further development of this approach.  

Another considered aspect is the relation between the parameters of an ensemble model and the ROC AUC score, Figure~\ref{parametrs}(a). The metric value increased with incrimination of the minimal number of samples required to be at leaf node (from 1 to 3) and after reaching a maximum it finally started to decrease. The enhancement of the minimal number of samples, required to split an internal node, leads to negligible growth of the score.

We have revealed the feature importance for a task of abbreviation detection which is shown in Figure~\ref{feature_importance}(a). The percentage of capital letters made the greatest contribution, while the other features (e.g., based on the presence of special characters and digits in the token) were palpably less significant. 

\subsection{Abbreviation identification}

At the second stage of experiments, it was assumed that abbreviation detection occurs using an ensemble of SVM, Random Forest, and Gradient Boosting. 

Two different classification models (Random Forest and Gradient Boosting) were used in this stage of experiments. As baselines, two different algorithms by~\cite{Schwartz_and_Hearst} and~\cite{glass2017language} were applied. Performance comparison on the abbreviation identification task is shown in Table~\ref{tab:identification_result}. The proposed algorithm showed a great improvement of the ROC AUC score and a perceptible improvement of two other metrics. This displays a promising outlook of approach, based on two consecutive classifications, to the abbreviation identification task.   

The performance of the Random Forest achieved ROC AUC score \textbf{0.926}, accuracy \textbf{0.994}, and F1-score \textbf{0.706}. The algorithm found 772 pairs of abbreviations and definitions, in which 7 were correct. The classifier marked 6 of them properly and overlooked 1 correct pair. Also, it marked 4 incorrect pairs as correct. 

Additionally, we have explored the relations between parameters of the Random Forest and the ROC AUC score, Figure~\ref{parametrs}(b). Experiments showed a strong dependence of ROC AUC on the minimal number of samples required to be at the leaf node. The score kept approximately the same in the case of parameter values from 1 to 7 and dramatically decreased starting from the parameter value 8.

\begin{figure}[t]
\centerline{\includegraphics[scale=0.2]{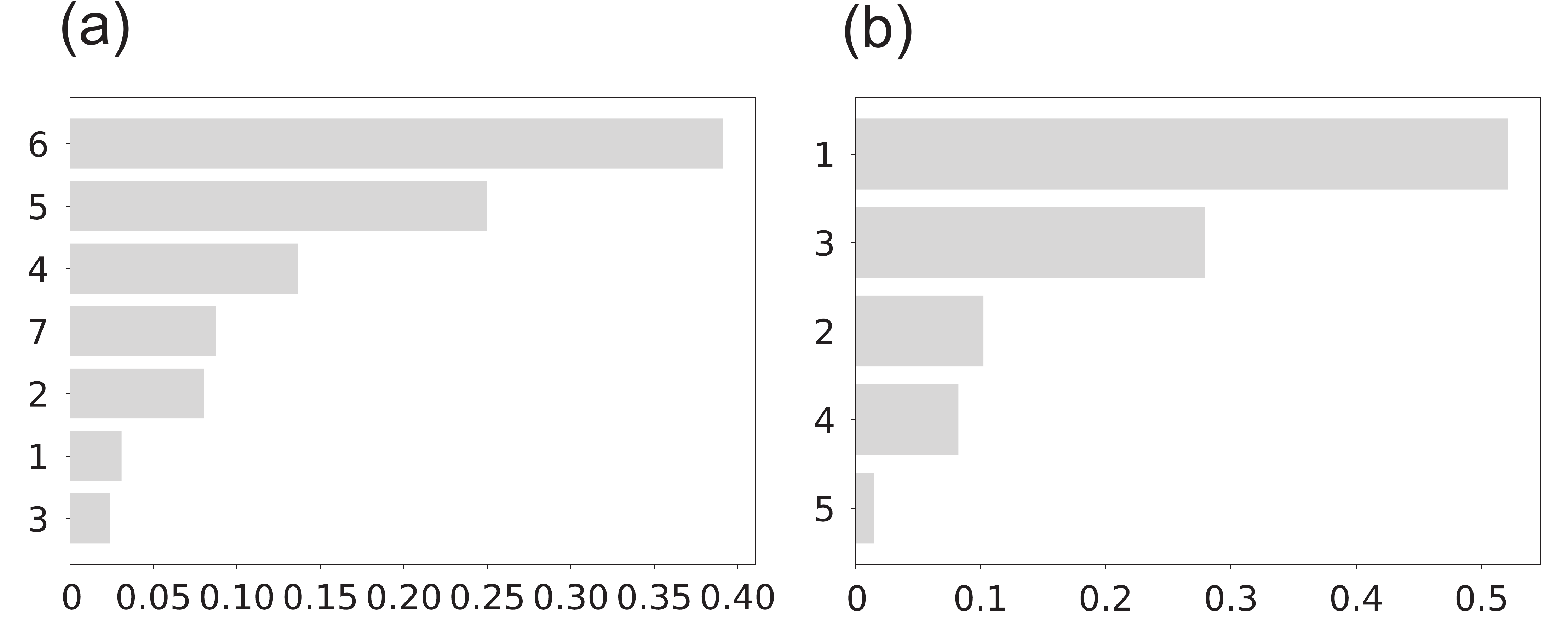}}
\caption{(a) Abbreviation detection feature importance:  (1) special characters presence feature, (2) letters and digits presence, (3) vowels and consonants feature, (4) magnitude of a length of the token, (5) presence of capital letters inside the word importance, (6) percentage of capital letters, and (7) presence of a token in a dictionary. (b) Abbreviation identification features importance: (1) distance between the abbreviation and the definition in the text, (2) number of letters of the abbreviation that match the first letters of tokens from the definition, (3) presence of an abbreviation of a definition inside the parenthesis feature, (4) semantic similarity between abbreviation and definition, (5) longest common subsequence of the abbreviation and the definition feature.}
\label{feature_importance}
\end{figure}

The feature importance is shown in Figure~\ref{feature_importance}(b). Distance between an abbreviation and identification was the most significant feature, and the longest common subsequent feature contributed almost nothing.

\section{Conclusion and Future work}

In this work, we considered the abbreviation identification task. A new approach to solve the problem of recognizing atypical abbreviations in Russian medical texts was proposed. First, we constructed the machine learning-based algorithm. Second, a set of rules to generate possible definitions were introduced. Third, we developed an algorithm for abbreviation identification. Experiments confirmed the efficiency of the proposed pipeline.

Besides, during this study, we assembled a new dataset for the task of abbreviation identification in Russian medical text. In further studies, we intend to broaden the task to cover cases in which the correct definition is not presented in the text.

\section*{Acknowledgment}
We acknowledge the contribution of Roman Senchenko and Vadim Porvatov.

\bibliographystyle{IEEEtran}
\bibliography{references.bib}
\end{document}